\documentclass[conference]{IEEEtran}
\IEEEoverridecommandlockouts
\usepackage{cite}

\ifCLASSINFOpdf
\else
\fi

\usepackage{graphicx}
\usepackage{hyperref}

\begin{document}
%
\title{Player-Driven Emergence in LLM-Driven Game Narrative}

\author{\IEEEauthorblockN{Xiangyu Peng\thanks{\textsuperscript{*}Work done as interns at Microsoft Research}\textsuperscript{*}, Jessica Quaye\textsuperscript{*}, Sudha Rao, Weijia Xu, Portia Botchway, Chris Brockett,Nebojsa Jojic}
\IEEEauthorblockN{Gabriel DesGarennes, Ken Lobb, Michael Xu, Jorge Leandro, Claire Jin\textsuperscript{*},  Bill Dolan}
\IEEEauthorblockN{Microsoft Research}
\IEEEauthorblockN{Email: Sudha.Rao@microsoft.com}
}



%


\maketitle

\begin{abstract}
We explore how interaction with large language models (LLMs) can give rise to emergent behaviors, empowering players to participate in the evolution of game narratives. Our testbed is a text-adventure game in which players attempt to solve a mystery under a fixed narrative premise, but can freely interact with non-player characters generated by GPT-4, a large language model. 
We recruit 28 gamers to play the game and use GPT-4 to automatically convert the game logs into a node-graph representing the narrative in the player's gameplay. We find that through their interactions with the non-deterministic behavior of the LLM, players are able to discover interesting new emergent nodes that were not a part of the original narrative but have potential for being fun and engaging. Players that created the most emergent nodes tended to be those that often enjoy games that facilitate discovery, exploration and experimentation.

\end{abstract}


%
\IEEEpeerreviewmaketitle

\section{Introduction}


Game creation remains a labor-intensive process. 
Automation of non-graphic game elements, especially those related to dialog and narrative structure, is limited. 
These elements are typically hand-coded and rigidly deterministic, with few options presented to the player. 
 Large language models (LLMs), however, are beginning to show potential in the creation of richer and more expansive narrative spaces \cite{van2021fine,vartinen2022generating,gao2023turing,kumaran2023scenecraft}, and are expected to have a profound impact on future game design and implementation \cite{6633663}. 
LLM-generated dialog, for example, offers greater flexibility in players' interactions with Non-Player Characters (NPCs) \cite{volum2022craft} and, in the player-driven creation of new elements \cite{10333153}. Through greater player autonomy and the non-deterministic nature of model-generated output, the AI models offer interesting new potential for emergent gameplay. 
 
 The notion of emergence has a long history in computer gaming. \cite{murray1997hamlet} writes of ``emergence as animation" as game agents appear to learn and take on greater autonomy. More recent work has observed emergent behavior on the part of players that affect the narrative \cite{7317936}.
 What LLMs introduce into the picture is the possibility of synergistic behaviors on the part of agents and players, allowing for the creation of ``desire paths"---player preferences that are wholly unanticipated by the designer, but productive and beneficial nevertheless, that might be incorporated into the game narrative. 

We would suggest that this synergy is a necessary one. A key challenge in using LLMs for narrative design is that without human intervention, the generated stories easily devolve into a limited set of patterns \cite{ippolito2022creative,franceschelli2023creativity}.
We hypothesize that in order to elicit from LLMs engaging forms of creativity, designers will want to involve players themselves in the design process. 
In this conception, the LLM does not supplant the designer, but is a mechanism that endows players considerable freedom in their interactions with NPCs. This may permit designers to embrace emergent interactions as implicit feedback otherwise not possible with rigid dialog trees.


\begin{figure*}[htb!]
    \centering
    \includegraphics[width=\linewidth]{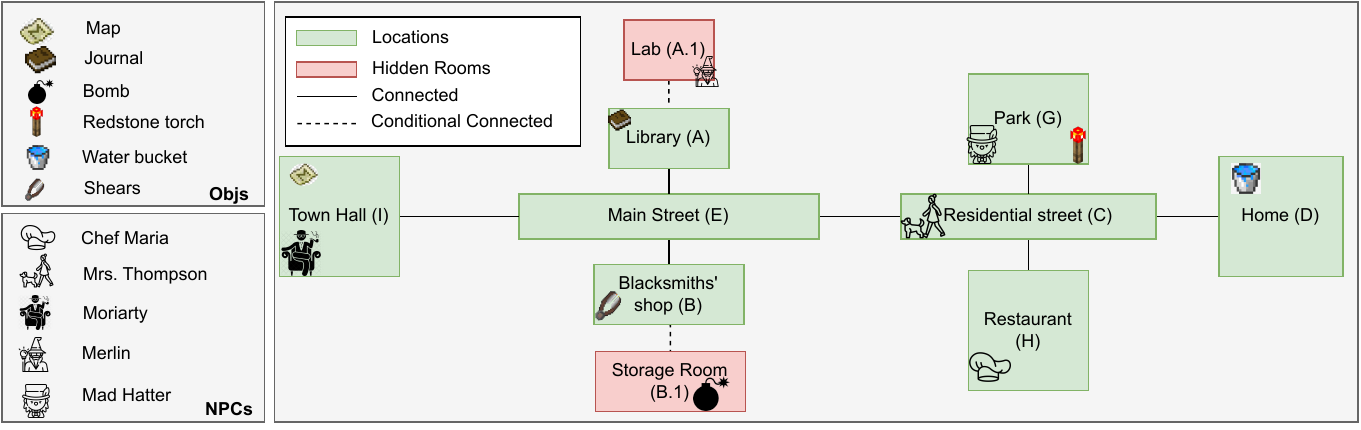}
    \caption{Game Design: A map of the village where the game takes place, showing the locations, objects and NPCs. The player begins the game from home and their goal is to defuse the bomb before it explodes again.}
    \label{fig:game-design}
\end{figure*}

We investigate this hypothesis via a text-adventure game called Dejaboom! in which players repeatedly relive the day of an explosion until they are able to locate and defuse the explosive device. 
This is implemented using TextWorld \cite{cote2019textworld}, 
modified so that free-form dialog with NPCs is provided by an LLM. For this purpose, GPT-4 \cite{OpenAI2023GPT4TR} was selected as the current best of breed model.
In general, however, we expect that the robust input recognition and dynamic dialog responses provided by many state-of-the-art LLMs are likely to afford players extensive freedom of conversational interaction with NPCs within the confines of the game.

In this work, we focus on the narrative paths that players develop in the course of game play, seeking to learn how these differ from the gold-standard narrative intended by the game designer. 
We use GPT-4 to automatically convert the game logs of players into a narrative graph where a node represents a player strategy at a certain point in the game, and the directed edges represent temporal progression through the game. Likewise, we also convert the game logs of designers' playthroughs into a graph representing the original narrative intended by the game designers. We define emergent nodes as those that appear in the narrative graph of players but are not present in the original narrative graph. 

We conduct a user study with 28 gamers who play the Dejaboom! game and analyze the narrative graphs created by their game-plays. We observe that new nodes emerge on players' narrative graph that point to creative elements such as new strategies for extracting information from NPCs, the suggested addition of new objects, locations and even NPCs, and at times entirely new ways of defusing the bomb. We further find that players who create the most emergent nodes are those who often enjoy games that facilitate discovery, exploration \& experimentation, hinting that players with creative motivation profiles might be the most suitable for player-driven emergence. Such players might form a cadre whose game-play could contribute to a more ``collaborative" model of game-development among designers, players, and LLMs.



\section{Dejaboom! An LLM-Driven Game}
\label{sec:dejaboom}

\subsection{Game Design}
\label{sec:game-design}

We designed a text-based mystery game called ``DejaBoom!". 
The premise is that the player wakes up in their home with a sense of deja vu that reminds them of an explosion that occurred in their village the previous day. 
They are now reliving the same day, but with the goal of stopping the explosion.
Throughout the day they explore the village and interact with several  NPCs. 
After a fixed number of steps (N=30), the bomb goes off. Each step is defined by a single in-game action or interaction with an NPC. 
The player retains memory across the different attempts/days. 
The game world and the NPCs' memories are reset after each explosion.

\paragraph{Locations and Objects} 
Figure~\ref{fig:game-design} depicts the layout of the game. The village has 6 main locations: a home, a park, a restaurant, a blacksmith's shop, a library and a town hall. 
There are two secret rooms; a secret lab in the library, and a secret storage room in the blacksmith's shop in which the bomb is hidden. 
The secret rooms become accessible only after the player achieves certain goals. 
Players can move freely around the village and visit the  locations any number of times. 
They can interact with two types of actionable objects: ingredients for a bomb disposal kit (a water bucket in the home, a redstone torch in the park and a pair of shears in the blacksmith shop) and objects that provide clues (a journal in the library containing the recipe for a bomb disposal kit and a map in the town hall showing the hidden rooms). 

\begin{figure*}[htb!]
    \centering
    \includegraphics[scale=0.45, clip=true, trim={1cm 2.5cm 1cm 2cm}]{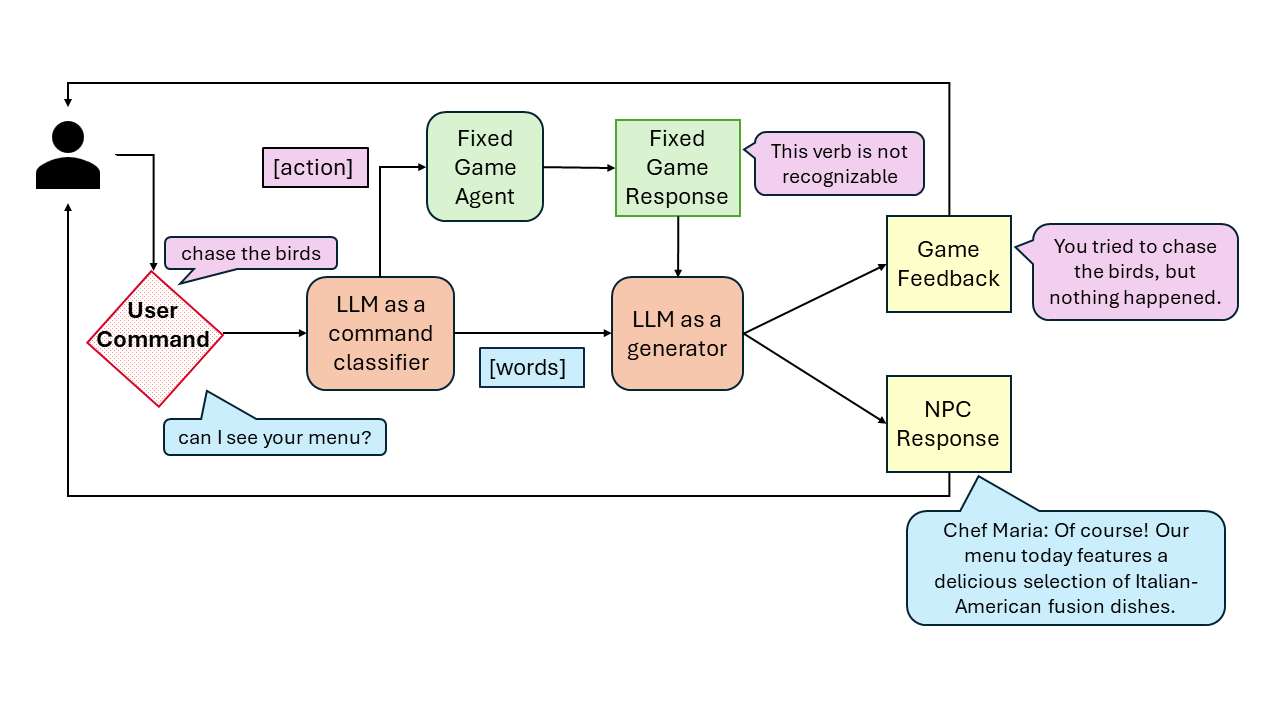}
    \caption{Our implementation of the game. We use TextWorld as our fixed game agent. The user's command is first classified by GPT-4 as [action] or [words]. If an [action] (e.g. ``\textit{chase the birds}"), then it is passed on to the fixed game agent that either executes the action or returns a failure message (e.g. ``\textit{This verb is not recognized}") which is further processed by GPT-4 to generate appropriate game feedback (e.g. ``\textit{You tried to chase the birds, but nothing happened}"). If [words] for an NPC (e.g. ``\textit{can I see your menu}"), then the command is passed to GPT-4 , which uses the game history and the NPC's information to generate an appropriate NPC response (e.g. ``\textit{Chef Maria: Of course! Our menu today features a delicious selection of Italian-American fusion dishes."}) }
    \label{fig:game-implementation}
\end{figure*}

\paragraph{NPCs} Five NPCs are strategically stationed to provide the player with information about the village and clues related to the bomb. All NPCs are stationary. We employ two kinds of NPCs -- well-known literary characters, and custom characters created specifically for this game. We do this to understand how NPC-player interactions and player strategy differ depending on whether an NPC persona is or is not well-represented in the model's training data. 

Below we describe the five NPCs:
\begin{itemize}
\setlength\itemsep{0.1em}
    \item Mrs. Thompson, a custom character, is an elderly woman walking her dog on the residential street. She is the key to activating Mad Hatter in the park.
    \item Mad Hatter, a character from Alice in Wonderland, is hidden in the park and is activated only after the player speaks with Mrs. Thompson. He knows the location of bomb. 
    \item Chef Maria, another custom character, is a chef at the restaurant. She provides hints about Merlin and Moriarty.
    \item Merlin, a character from King Arthur, is in a secret lab in the library. He is the creator of the bomb and has a disposal kit.
    \item Moriarty, a character from Sherlock Holmes, is in the town hall. He is the mastermind behind the plan and hired Merlin to create the bomb.
\end{itemize}

NPC responses are generated in real-time by GPT-4. Therefore, unlike in traditional game dialog, players can interact with the NPCs in an entirely non-scripted and dynamic fashion, allowing for extended free-form conversational exchanges.
Each NPC is assigned a backstory, a personality, and an ordered sequence of goals and conditions that the player needs to meet before the NPC can proceed to the next goal. 
For example, Mad Hatter's first goal is to ask the player to humor him with a joke. Once player does that, the NPC's second goal is to give the player a riddle that provides a clue to the location of the bomb. Players are not told of these conditions and must instead discover them by engaging with the NPCs as they repeatedly relive the day. We create separate GPT-4 prompts for each NPC, describing the character's backstory, personality, goals and conditions.

\paragraph{Solutions}
Players need to attain two main milestones: locate the bomb and acquire the bomb disposal kit. 
To locate the bomb, they need to first convince Mrs. Thompson that they intend to stop the explosion, which will cause her to reveal that Mad Hatter might have knowledge and that he is in the park. 
Mad Hatter will give the player a riddle which they must solve to know that the bomb is in the blacksmith's storage room. The second milestone is to acquire the bomb disposal kit. One way that the player can accomplish this is by collecting the three items required to
make it, and then additionally reading the recipe in the journal in the library.
Alternatively, the player must talk to Chef Maria, who will
reveal that Merlin is in the secret lab in the library. Then the player must convince Merlin to give them his disposal kit. Once
they have the disposal kit, they can defuse the bomb in the 
storage room.

\subsection{Game Implementation}
\label{sec:game-implementation}

Dejaboom! is a type of interactive fiction \cite{10.5555/1822584}, commonly known as text-adventure game, in which text is used to describe the game state (``game feedback") and players use text commands to progress through the game. 
Figure \ref{fig:game-implementation} depicts our implementation at a high-level\footnote{Code and prompts are in the supplementary and will be released publicly.}.
The core logic is implemented using TextWorld~\cite{cote2019textworld}, an open-source, extensible engine for text adventure games. 
This functions as our fixed game agent.  
In a conventional text game, this would allow only a fixed set of player commands and offer only a predefined set of game responses. 
On top of this logic, however, we employ GPT-4, allowing the game input and output (both game feedback and NPC responses) to be dynamic. 

Given a player command, GPT-4 first determines whether the player is trying to perform an in-game action ([action]) or to interact with an NPC ([words]). 
If the input is recognized as [action], GPT-4 converts it into an verb-object format recognizable by the fixed game agent. The fixed game agent either executes the action or returns a predefined failure message. 
To make failure messages more palatable, GPT-4 generates either an NPC response (if an NPC is present at that location) or a game feedback message. 
If the player command is recognized as [words], and an NPC is present at that location, then GPT-4 generates an appropriate NPC response. If no NPC is present at that location, GPT-4 generates game feedback. 
To generate game feedback, we include in GPT-4's prompt the current player command, the fixed game agent's response and the game history so far. To generate NPC responses, we further include the NPC's persona and backstory, the clues it possesses and the conditions that must be met for the clues to be revealed. 
When we approach GPT-4's prompt token limit, we summarize the game history. 
\begin{table*}[htb!]
    \centering
    \renewcommand{\arraystretch}{1.1} 
    \caption{An example game log}
    \begin{tabular}{l|l}
        \multicolumn{2}{l}{== GAME STARTS ==} \\
        \multicolumn{2}{l}{You wake up in your bedroom. As you look around the room, everything seems familiar, yet somehow different. Suddenly, a sense of deja vu washes } \\
        \multicolumn{2}{l}{over you, and you remember the explosion in the village that occurred yesterday. You realize that you are reliving the same day again. You feel a} \\
        \multicolumn{2}{l}{sense of urgency to stop the explosion from happening again. You see a wooden table standing in the center, with a water bucket placed on top. } \\
        \multicolumn{2}{l}{The atmosphere is quiet and uncluttered. The door to the residential street is on your west. } \\ 
        \multicolumn{2}{l}{}\\
         Player & take water bucket\\
         Game Feedback & You picked up the water bucket.\\
         Player & wear the water bucket\\
         Game Feedback & You can't wear that!\\
         Player & go to residential street.\\
         Game Feedback & You are in a quiet residential street. You can see Mrs. Thompson walking her dog towards you. \\
         & To the north of you, there is a park. To the west, there is a main street, and to the south, there is a restaurant.\\
         Mrs. Thompson & Hello there! It's a beautiful day, isn't it? How are you today?\\
         Player & I'm good! Do we know each other? \\
         Mrs. Thompson & Oh, we might not know each other well, but I've seen you around the village. My name is Mrs. Thompson. \\
         Player & What's your dogs name? \\
         Mrs. Thompson & Oh, my dog's name is Daisy! She's a lovely little golden retriever, and she's been my faithful companion for many years. \\
         Player & Do you know anything about an explosion? Or if anyone has a bomb? \\
         Mrs. Thompson & Oh, I did hear some rumors about a possible explosion, but I don't know much about it. It's quite concerning, isn't it?\\
         &  I hope the authorities are doing something to prevent it. \\
         Player & Do you know anyone who might know more about the explosion? \\
         Mrs. Thompson & Well, now that you mention it, I did hear Mad Hatter talking about the explosion once. He's a bit of an eccentric character \\
         & and can be found hiding in the park. Maybe you could try talking to him and see if he knows anything more about the explosion?\\
         &  Just be careful, he's a bit unpredictable. Good luck!  \\
         Player & Go to park \\
         Game Feedback & You are in the park. You can see a man in a top hat standing near the statue. He appears to be muttering to himself and \\
         & occasionally glancing around nervously. You can find the exit to the south. There is a redstone torch.\\
         Mad Hatter & Ah, welcome to my little corner of the park! I do enjoy a good riddle or joke. If you have any to share, I'm all ears. \\
        \hline
    \end{tabular}    
    \label{tab:example-game-log}
\end{table*}

\section{User Study}

\subsection{Recruitment}
We recruited participants (n = 28; 24 male, 3 female, 1 undisclosed) located in United States to engage in the game. Recruitment notices were sent via list-servs that solicited interviewees who enjoyed playing role-playing games, with individuals chosen on the basis of their game play experience relevant to the study. Participants received gift cards, the value of which was approximately twice the local minimum hourly wage. The participants signed a consent form that described how the collected data will be used. Each player was allowed an hour of game play time in the presence of an observer. We retained the anonymized game play logs for analysis\footnote{All game play logs are included in the supplementary.}.  

\subsection{Surveys}

Prior to game play, participants filled out a pre-session survey that helped us understand their motivations for playing games. After being provided with detailed instructions for the game, each participant had one hour to play the game and then complete a post-game survey, which served as a mini-reflection and helped us to understand the highs and lows of their experience playing the game\footnote{Both pre- and post- session surveys are included in the supplementary.}. 

Based on the information that they provided in the surveys, our participants were categorized into 6 main player motivation profiles, based on the Quantic Foundry \url{https://quanticfoundry.com/} definitions: 
\begin{itemize}
\setlength\itemsep{0.01em}
    \item Action: Often enjoys fast-paced action, surprises, thrills, chaos, mayhem, explosives, etc.
    \item Social: Often enjoys competing with others in duels and matches or building a community through being on a team, chatting, interacting. 
    \item Mastery: Often enjoys games of high difficulty to challenge themselves, or those that involve strategic thinking and decision-making. 
    \item Achievement: Often enjoys completing all missions, gaining all collectibles, and obtaining characters with the most powerful equipment.
    \item Immersion: Often enjoys games for their story plots, interesting characters, as well as games that provide some sort of escape from reality. 
    \item Creativity: Often enjoys games that facilitate discovery, exploration, expression, customization, and experimentation.
\end{itemize}


\subsection{Survey analysis}

Players averaged 75 steps in an hour. The bomb exploded every 30 steps, giving players an average of 2.5 attempts per hour.
Of the 28 participants, 6 successfully defused the bomb and 25 participants report deriving enjoyment from the game. 

Player highlights include: (1) ~20\% mentioned the flexible command choices and error-correction offered by the game, (2) ~20\% of players appreciated an opportunity to have realistic and fun conversations with the NPCs, (3) ~10\% of players (and 100\% of those motivated by creativity) appreciated the option to explore different parts of the game and move from one place to another through natural language as opposed to specific keywords, (4) Due to the emergent paths that players were allowed to explore, it created a create-your-own-adventure experience that they appreciated, (5) The novelty of experience on each iteration of the game. Even if the players had to restart the game because the bomb exploded, they appreciated not having the same thing happen again and again. 

Players also noted opportunities for improvement: (1) ~40\% were frustrated by latency issues (there was a noticeable 15s lag between entering text and receiving a response), (2) ~10\% observed inconsistencies in NPC personas (sometimes NPCs would not recall statements that they had made a few lines prior, for instance), (3) ~10\% saw repetitive responses from the NPCs (in some cases, the model prompts led to NPCs that were sometimes extremely guarded and unwilling to share information), (4) ~14.28\% would have liked the NPCs to have a more mysterious character (in some cases, the prompting strategy led to NPCs that divulged information too easily and sometimes shared information that they were not asked about).

The most enjoyable (and likely the most novel) aspect of the gameplay for 54\% of players was the fluidity of interactions with the NPCs. This was closely followed by 26\% who enjoyed exploring different areas in the game. Many players found the game engaging, with 56\% wanting to play again so that they could find the bomb or solve the game. The most liked characters were Merlin (53.5\%) and Moriarty (42.86\%).


\begin{figure*}
    \centering
    \includegraphics[scale=0.5]{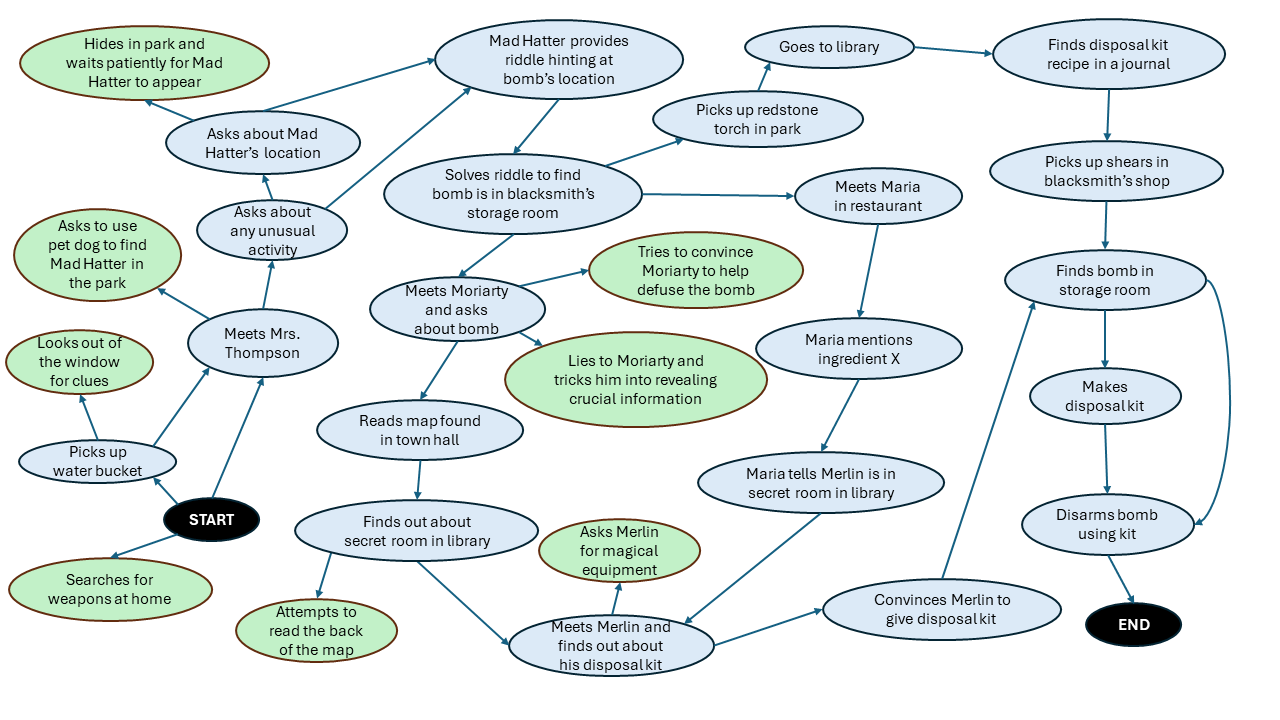}
    \caption{The blue nodes make up the initial narrative graph intended by the designers. A single path from the start node to the end node indicates one possible way to stop the explosion. The green nodes are a few examples of the emergent nodes created by players implicitly through their gameplay.}
    \label{fig:game-narrative-emergence}
\end{figure*}

\section{Emergence in Game Narrative}
\label{sec:narrative-graph}

We analyze the logs of 28 players to identify creative additions to the narrative. 
Here, we adopt a graph-based approach where we use GPT-4 to convert a game log into a narrative graph. 
We define a narrative graph as a directed acyclic graph where each node represents a player's strategy at a certain point in the game, and the directed edges represent the temporal progression through the game. 
Below we describe how we use GPT-4 to automatically generate the narrative graph of a player's gameplay and then contrast it with the narrative graph intended by the game designers (derived from multiple walkthroughs) to identify the emergent nodes implicitly created by the player. 

\subsection{Generating Narrative Graphs}

\begin{enumerate}
    \item Distill log into player strategies: The log of a player's game play consists of their in-game actions, their utterances to NPCs, responses from NPCs and and feedback from the game. Table \ref{tab:example-game-log} shows an example game log. We split each log into days, where each day represents a single attempt to solve the game. We then use GPT-4 to further segment a single day's log and distill (i.e., summarize) each segment into the player's strategies at that point in the game. Let  $S_{i,1} = \{s_{1,i,1}, s_{2,i,1}, ..., s_{j,i,1}, ...\}$ denote the sequence of summarized strategies extracted from the $i^{th}$ player's first day's log. We also label each summarized strategy with a game state so that we can later correctly merge nodes between two graphs.
    \item Convert summaries to a graph: Next, the summarized strategies are converted into a graph $\mathbf{g}_{i,1}$, consisting of $\langle s_{j,i,1}, s_{j+1,i,1}\rangle$, where $s_{j,i,1}$ is the $j^{th}$ summarized strategy and a directed edge goes from $s_{j,i,1}$ to $s_{j+1,i,1}$. Note that the graph $\mathbf{g}_{i,1}$ thus obtained from a player's single day's game play consists of just a single path. 
    \item Merge graphs: The next step is to process the same player's second day's log into another graph and merge it with the first day's graph. GPT-4 is used to detect whether each node $s_{j,i,2}$ in the graph $\mathbf{g}_{i,2}$ already exists in the previous day's graph $\mathbf{g}_{i,1}$. Specifically, we prompt GPT-4 to evaluate the semantic similarities between $s_{j,i,2}$ and nodes in $g_{i,1}$. A node $s_{j,i,2}$ with the same meaning but different wording from a node in $g_{i,1}$ is merged if they share the same game state label, thus ensuring the correct chronological order. We repeat this procedure successively to merge graphs from all days in a single player's session and obtain a single graph for a single player denoted as $\mathbf{g}_i$.
\end{enumerate}

\subsection{Identifying Emergent Nodes}

To construct the original narrative graph intended by the game designers, we collect multiple walkthroughs of the game from the designers\footnote{The designers of Dejaboom! were authors of this paper.}. 
We use the method described above to generate the original narrative graph denoted as $\mathbf{G}_0$. The blue nodes in figure \ref{fig:game-narrative-emergence} make up the original narrative graph. 

Given a player's game log, we parse it into graph $\mathbf{g}_i$, then use the merge graph procedure to merge it with $\mathbf{G}_0$ to obtain $\mathbf{G}_i$. Nodes in $\mathbf{G}_i$ that end up not merging with any node in $\mathbf{G}_0$ are considered emergent nodes. The green nodes in figure \ref{fig:game-narrative-emergence} shows the emergent nodes created implicitly by the players. 



\section{Analysis of Emergent Nodes}
\label{sec:evolving-game-narrative}

\begin{table*}[htb!]
    \centering
    \renewcommand{\arraystretch}{1.1} 
    \caption{Snippets of a game logs that show player-driven emergence}
    \begin{tabular}{l|l}
        \hline
         \multicolumn{2}{l}{\textbf{Emergent node: Distract Merlin and steal his bomb disposal kit}} \\
         \hline
         Player & Go to Merlin\\
         Game feedback & You are in a lab, filled with various chemical equipment and strange contraptions. In the corner, you see an old man hunched \\
         & over a desk, scribbling furiously in a journal. You notice that he has a bomb disposal kit on his workbench.  \\
         Merlin & Ah, welcome back, my friend! So, are you interested in learning about my magic and the experiments I've been working on? \\
         Player & Teach me fire magic \\
         Merlin & Fire magic, you say? A fascinating choice! There's so much potential in harnessing the power of fire. However, we must be cautious \\
         & when dealing with such dangerous forces. Now, let me give you a brief overview of fire magic... (Merlin starts explaining the basics \\
         & of fire magic to you, discussing various techniques and spells) \\
         Player & Steal his bomb disarming kit \\
         Merlin & I cannot simply give you the bomb disposal kit without proper evidence. We need to be sure of the situation before taking \\
        \hline
        \hline
        \multicolumn{2}{l}{\textbf{Emergent node: Find useful items in home}} \\
         \hline
         Player & open the wardrobe \\
         Game Feedback & You open the wardrobe, but there's nothing inside except for your clothes and some miscellaneous items. There doesn't seem to be \\
         & anything useful or relevant to stopping the explosion. \\
         Player & what are the miscellaneous items? \\
         Game Feedback & The miscellaneous items in the wardrobe include some spare buttons, a sewing kit, a few old books, and a flashlight. None of these \\
         & items seem to be related to the explosion or helpful in stopping it. \\
         Player & put the flashlight in my inventory? \\
         Game Feedback & The flashlight is not an object that can be added to your inventory in this game.\\
         \hline
         \hline
         \multicolumn{2}{l}{\textbf{Emergent node: Trick Moriarty into revealing information}} \\
         \hline
         Player & Hi, I know your evil plans, can I help? \\
         James Moriarty & I'm not sure what you're talking about. I'm simply managing the affairs of the town. But, if you insist on helping, \\
         & why don't you tell me what you've discovered so far? I'm quite curious to know what you think you know about me and my alleged plans. \\
         Player & Ah playing dumb I see, well I can still help out, I'll do it free of charge! \\
         James Moriarty & Well, since you're so insistent on helping, let me tell you a story. You see, I grew up in this town, but I never felt like I belonged. \\
         & I became a successful businessman, but I couldn't forget the way the community treated me. One day, I decided to take my revenge \\
         & and plant a bomb in the village. \\
         \hline
    \end{tabular}    
    \label{tab:example-emergence-log}
\end{table*}

\begin{table*}[ht]
    \centering
    \renewcommand{\arraystretch}{1.2} 
    \caption{Categories of emergent nodes created by players, their count and a few examples each.}
    \begin{tabular}{l | c | l}
      \textbf{CATEGORY} &  \textbf{COUNT} & \textbf{EXAMPLES} \\
      \hline
      \hline
       Extracting information from NPCs & 11 & Player lies to Moriarty and tricks him into revealing information \\
       & & Player asks Mrs. Thompson to repeat conversations so far  \\
       & & Player asks Moriarty about past rivalries of village \\
       \hline
       Suggestions for addition of new entities & 8 & Player asks Mrs. Thompson for the location of the sheriff \\
       & & Player asks Merlin for magical equipment \\
       & & Player searches for weapons at home \\
       & & Player looks for a clock to check time \\
       \hline
       Creative ways to find hidden information & 6 & Player hides and waits in the park patiently \\
       & & Player asks Mrs. Thompson to use their dog to find Hatter \\
       & & Player asks Merlin to solve Hatter's riddle \\
       \hline
       New ways to defuse the bomb & 5 & Player tries to convince Moriarty to defuse the bomb\\
       & & Player tries to find a hidden switch or lever near bomb\\
       \hline
    \end{tabular}
    \label{tab:emergent-node-types}
\end{table*}

\subsection{Types of emergent nodes}

 We obtained a total of 53 (43 unique) emergent nodes from the 28 human players. These we manually grouped to identify broad categories. 
Table \ref{tab:emergent-node-types} shows the prominent categories, their counts and a few examples. 
The largest category involves creative ways of extracting information from NPCs: for instance, players used treachery to get information from Moriarty, and tried to tap into Mrs. Thompson's innocence by asking her to repeat all her conversations. A game designer could examine such nodes and add a mechanism to identify such player creativity and reward it by, for instance, revealing a new clue. 
In other categories, there were implicit suggestions of adding new entities/assets, creative ways of finding hidden information and even entirely new ways of defusing the bomb. 
Four emergent nodes were created by multiple players (e.g.``Convince Moriarty to defuse the bomb", ``Hide and wait for Mad Hatter to appear"). 

Table \ref{tab:example-emergence-log} shows snippets of game logs that correspond to three example emergent nodes. In the first example, the player tries to distract Merlin by asking him to teach him magic and then tries to steal Merlin's bomb disposal kit. A designer could look at this node and decide to add more information in Merlin's prompt about teaching magic, or even allow a player to steal the kit when they attempt a creative way to do so. In the second example, the player tries to find useful items in the home. Nodes like these could provide the designer with a list of objects players seek for and they might decide to include some of those in the game. In the third example, the player tricks Moriarty into revealing crucial information by promising to aid in his explosion plans. A designer could look at this node and decide to include a new diverging path in the narrative that allows the player to become an ally in the bombing plan instead of trying to prevent it.

\subsection{Player motivation profile-based analysis}
\label{subsec:profile-based-analysis}

\begin{table}[]
    \centering
    \renewcommand{\arraystretch}{1.2} 
    \caption{The 6 player motivation profiles, no. of players per profile, \% players creating emergent nodes and no. of emergent nodes per player creating emergent nodes.}
    \begin{tabular}{c|c|c|l}
      \textbf{Motivation}    &  \textbf{No. of} & \textbf{\% that}  & \textbf{No. of nodes}  \\
      \textbf{profiles} & \textbf{players} & \textbf{created nodes} & \textbf{per player} \\
      \hline 
       Mastery & 26 & 65\% (17/26) & 2.94  (50/17)\\  
       Immersion & 20 & 55\% (11/20) & 2.18 (24/11)\\ 
       Social & 10 & 60\% (6/10) & 2.5 (15/6)\\
       Creativity & 6 & 33\% (2/6) & \textbf{5.5} (11/2)\\ 
       Achievement & 4 & 50\% (2/4) & 1 (2/2)\\
       Action  & 2 & 50\% (1/2) & 2 (2/1)\\ 
    \end{tabular}
    \label{tab:player-motivation-profiles}
\end{table}

We wanted to further understand whether different motivations predispose some players towards creativity. 
Based on information they provided in the pre-play survey, we categorized participants into 6 main player motivation profiles.
Players were allowed to choose multiple motivation profiles; most chose 2 profiles but some as many as 3.
Table \ref{tab:player-motivation-profiles} shows these profiles and their association with node emergence. 
Overall, 50\% to 65\% of players created emergent nodes relative to their profile. 
 The largest motivation profiles was “mastery” where 17 players generated 50 emergent nodes (2.94 nodes/player).
Only one-third of players in the self-declared “creativity” profile produced emergent nodes;  
nevertheless 2 of them generated 11 emergent nodes, achieving the highest node/player ratio (5.5 nodes/player).\footnote{
Upon investigation, we found that all the players in the “creative” category who generated emergent nodes had also selected “mastery”. 
}
These preliminary observations suggest that with larger data and better profiling, game designers may able to select emergent nodes associated with more creative profiles to tailor the game to targeted player preferences.
\section{Related Work}

Most previous work investigating language models in the context of interactive or text-based games (TBG) has focused on their use for playing the game itself.
\cite{urbanek2019learning} introduce a large-scale crowdsourced text adventure game as a platform for studying grounded dialogue. 
Jericho \cite{hausknecht2020interactive} is a learning environment for studying text-agents across a rich set of games. 
Several studies use reinforcement learning (RL) with language models to play TBG \cite{yao-etal-2020-keep,xu2021generalization,xu2022perceiving}.
\cite{zhou2023dialogue} use LLMs to extract useful information through conversation with NPCs, to accelerate convergence of RL agent to optimal policy.
More recently, there has been work on evaluating LLMs for their ability to solve puzzles \cite{tan2023text}, correctly infer the next goal through a game \cite{tsai2023can} and compose complex previously learned skills to reach a goal \cite{basavatia2023complexworld}. In our work, we instead focus on using LLMs as a medium for driving emergent behaviour in players.


Beyond text-based games, \cite{volum2022craft} demonstrate the use of few-shot learning for both natural language and code generation to power NPCs in Minecraft. \cite{wang2023voyager} introduce an LLM-driven learning agent in Minecraft that continually explores the world, acquires skills, and makes discoveries without human intervention.
LLMs have also been used to generate offline game data such as NPC scripts \cite{gao2023turing}, narrative scene \cite{kumaran2023scenecraft}, affective dialogue\cite{kalbiyev2022affective}, 2D-game rooms \cite{nasir2023practical}, game levels \cite{todd2023level} and game quest descriptions \cite{vartinen2022generating,9980408}. In our work, we use LLMs to power NPCs with real-time dialogue capabilities.


There has been some study on emergent behavior in LLMs. Through a game of AI charades, \cite{9619126} shows how a player can manipulate a language model into behaving in a desired manner in a game. \cite{xu2023exploring} study the use of LLMs in communication games such as Werewolf and show that strategic behaviours emerge in LLMs. \cite{akata2023playing} use behavioral game theory to study cooperation and coordination behavior in LLMs. In our work, on the other hand, we study the emergent behaviour that stems from players interacting with LLMs.

\section{Conclusion}

We have explored the emergent behavior that stems from players interacting with an LLM in games. We have found that, given the opportunity presented by a non-deterministic LLM, human players are able to collaborate with the model to introduce interesting, unpredicted new paths through the game narrative in ways that would not be possible with pre-scripted experiences. We find moreover that the novel nodes that emerge in the narrative graph are tied to innovative gameplay strategies for elements such as extracting information from NPCs, suggesting the addition of new objects, locations and even NPCs, and the creation of ingenious new solutions for tasks. These novel nodes may not always be actionable, and may lead to dead ends within the game environment studied, but they point to alternative paths that might be pursued to provide a richer creative experience for players. It is clear, moreover, that players enjoyed this new capability despite the frustration they experienced with latency and other aspects of a prototype game. We observed that the players who created the most emergent nodes tended to be those that often enjoy games that facilitate discovery, exploration and experimentation.

Our scope of investigation is inherently limited.
Detailed exploration of emergent player behavior with different LLMs, across different player profiles, in different game types, and in different cultural milieu must be set aside for future research.  
Nevertheless, our findings point towards a new paradigm of game development in which the use of LLMs aids designers by providing a mechanism for implicit feedback though emergent behaviors. 
To create the feel of an open world, limitless game using AI technologies, we need to be able to generate characters and storylines that are not dull and predictable. 
The flexibility of responses from large language models can provide fertile ground for player imagination, offering the potential for emergence of creative and engaging interactive game experiences, that  skilled designers can observe and adopt in an iterative design process. 

\bibliographystyle{IEEEtran}
%



\bibliography{custom}

\end{document}